\documentclass[runningheads]{llncs}

\usepackage[dvipsnames,table]{xcolor}
\usepackage{eccv}
\usepackage{color}
\definecolor{defaultcolor}{gray}{.9}
\newcommand{\default}[1]{\cellcolor{defaultcolor}{#1}}
\usepackage{multirow} 

\usepackage{bbding}

\usepackage{eccvabbrv}

\usepackage{graphicx}
\usepackage{booktabs}
\usepackage[accsupp]{axessibility}  

\usepackage{hyperref}

\makeatletter
\renewcommand*{\@fnsymbol}[1]{\ensuremath{\ifcase#1\or ^{(}$\Envelope$^{)} \or \dagger\or \ddagger\or
   \mathsection\or \mathparagraph\or \|\or **\or \dagger\dagger
   \or \ddagger\ddagger \else\@ctrerr\fi}}
\makeatother

\usepackage{orcidlink}

\begin{document}

\title{Projecting Points to Axes: Oriented \\Object Detection via Point-Axis Representation}

\titlerunning{Point-Axis Representation}

\author{Zeyang Zhao\inst{1} \and  Qilong Xue\inst{2} \and Yuhang He\inst{1} \and Yifan Bai\inst{2}\and Xing Wei\inst{2}$^{(}$\Envelope$^{)}$ \and Yihong Gong\inst{1}}

\authorrunning{Z. Zhao et al.}

\institute{Institute of Artificial Intelligence and Robotics, Xi'an Jiaotong University\\
\and School of Software Engineering, Xi'an Jiaotong University\\
$^{}$\Envelope$^{}$ Corresponding Author\\
\email{\{zeyang, xql1119, yfbai\}@stu.xjtu.edu.cn}\\
\email{\{heyuhang, weixing, ygong\}@mail.xjtu.edu.cn}\\
\url{https://PointAxis.github.io/}}

\maketitle
\vspace{-0.5cm}
\begin{figure*}[h]
	\centering
	\includegraphics[width=0.8\linewidth]{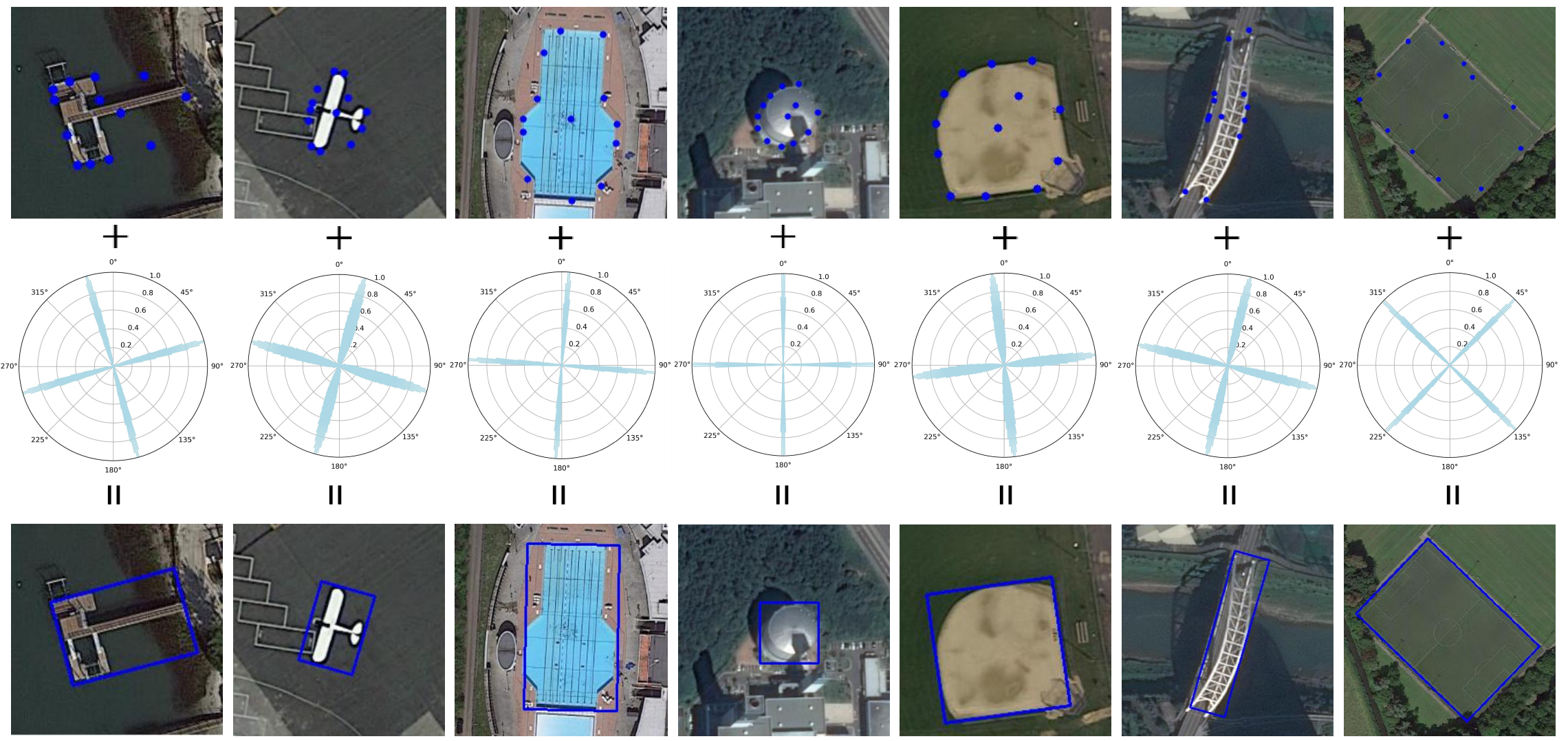}
	\caption{Robust oriented object detection results with point-axis representation.}
    \label{intro_vis}
\end{figure*}
\vspace{-1.2cm}

\begin{abstract}
This paper introduces the point-axis representation for oriented object detection, as depicted in aerial images in Figure~\ref{intro_vis}, emphasizing its flexibility and geometrically intuitive nature with two key components: points and axes. 1) \textbf{Points} delineate the spatial extent and contours of objects, providing detailed shape descriptions. 2) \textbf{Axes} define the primary directionalities of objects, providing essential orientation cues crucial for precise detection. The point-axis representation decouples location and rotation, addressing the loss discontinuity issues commonly encountered in traditional bounding box-based approaches. For effective optimization without introducing additional annotations, we propose the max-projection loss to supervise point set learning and the cross-axis loss for robust axis representation learning. Further, leveraging this representation, we present the Oriented DETR model, seamlessly integrating the DETR framework for precise point-axis prediction and end-to-end detection. Experimental results demonstrate significant performance improvements in oriented object detection tasks.

\end{abstract}

\begin{figure*}[t]
	\centering
	\includegraphics[width=0.95\linewidth]{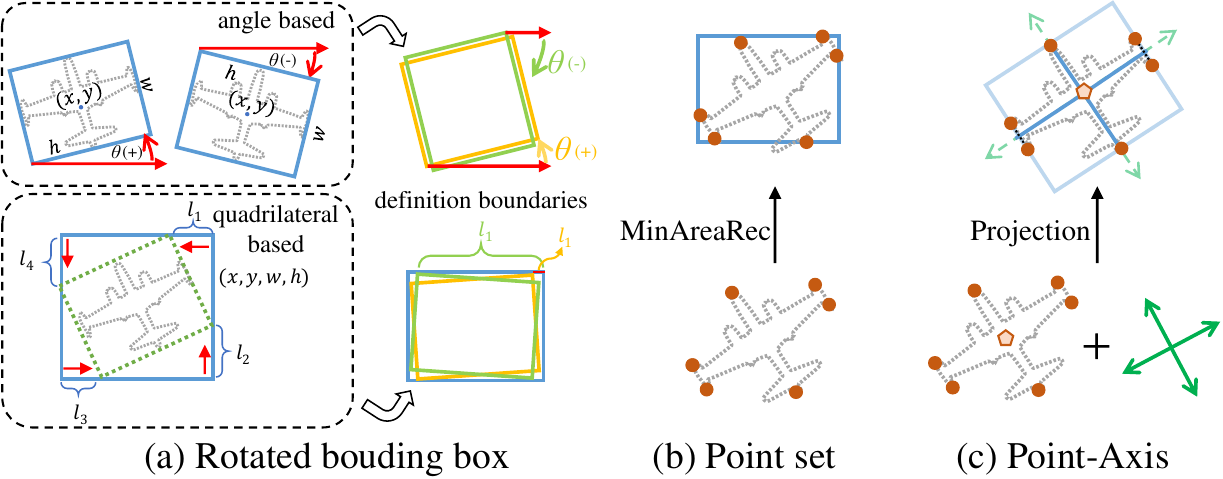}
	\caption{Mainstream oriented object representations. (a) is the \emph{rotated bounding box} representation, (b) is the \emph{point set} representation, and (c) is the \emph{point-axis} representation we propose.}
	\label{fig:intro_compare}
\end{figure*}

\section{Introduction}

Oriented object detection~\cite{dota,dotat,dior,mmrotate} is an important task in computer vision and has raised remarkable research interests in the community. Compared to vanilla object detection~\cite{girshick2015fast,redmon2016you,tian2019fcos,wang2023yolov7,zou2023object, song2024non} which only outputs the horizontal bounding box aligned with the image's axis, oriented object detection requires a more flexible and accurate approach to handle non-axis aligned objects.

Current methods for oriented object detection predominantly rely on rotated bounding boxes to represent objects, as illustrated in Figure~\ref{fig:intro_compare} (a). Despite variations in their definitions, these methods all encounter a common challenge known as loss discontinuity, stemming from abrupt changes in rotation degree or height-width definitions. Angle-based methods~\cite{roi,yang2020arbitrary,yang2022kfiou,yu2023phase} define a rotated bounding box as $({x,y,w,h,\theta})$, where $x$ and $y$ denote the center point location, $w$ and $h$ represent the width (short edge) and height (long edge), and $\theta$ denotes the orientation angle between the long edge of the box and the horizontal axis. 
Although these methods are effective in most cases, they suffer from a critical issue: when the lengths of the long and short edges are similar, the angle $\theta$ can switch between $\theta$ and $\theta \pm 90^{\circ}$, leading to discontinuities in the loss function. This switching behavior makes it difficult for the model to learn stable and consistent representations, thereby affecting detection performance.

Quadrilateral methods~\cite{rsdet,gliding,RIL}, on the other hand, determine a rotated bounding box using an outer horizontal box and offsets from its four corners, denoted $({x,y,w,h,l_1,l_2,l_3,l_4})$. While these methods provide more flexibility in representing objects' shapes and orientations, they also encounter ambiguous definitions. Specifically, when an object approaches a horizontal orientation, it encounters ambiguous definitions that the offset could change abruptly, as is the case shown in the bottom of Figure~\ref{fig:intro_compare} (a).
Other variant definitions, such as those based on box boundary vectors~\cite{bbvec}, middle lines~\cite{wei2020oriented}, or Gaussian distributions~\cite{gwd,kld}, also face various representation issues, for example, the square problem~\cite{psc_pami,theoretically}.

Recent research has drawn attention to object representations based on point set~\cite{reppoints,Li_2022_CVPR} as a promising approach to address these challenges. In these approaches, each object is represented as a collection of points. 
These methods then determine the object's bounding boxes by computing the minimum area rectangle enclosing these points. While these point set-based approaches excel at capturing the detailed locations of targets, their effectiveness heavily relies on the quality and quantity of the predicted points. Moreover, they often disregard the primary directionalities of objects, which can hinder their ability to accurately detect targets with complex shapes. In such cases, the computed rectangles may fail to adequately enclose the targets, as illustrated in Figure~\ref{fig:intro_compare} (b).

To overcome existing limitations, we present an innovative point-axis representation that offers a divide-and-conquer approach for describing oriented objects, as illustrated in Figure~\ref{fig:intro_compare} (c). Unlike methods that rely on direct orientation regression, our technique leverages a more flexible and geometrically intuitive framework centered on point sets and axes. 

This representation emphasizes two pivotal aspects.
\begin{itemize}
    \item \textbf{Points for Shape Descriptions}: The use of points to delineate the spatial extent and contours of objects provides a detailed shape representation. This aspect is crucial, especially when dealing with irregularly shaped objects that traditional bounding box-based methods might struggle to accurately describe.
    \item \textbf{Axes for Orientation Cues}: Beyond shape information, knowing the primary directionalities of objects is essential for precise detection. Axes provide these orientation cues, which are particularly important when keypoints can appear in near-circular shapes.
\end{itemize}
Our point-axis representation effectively disentangles location and rotation, facilitating a more flexible and continuous modeling approach for both aspects. Additionally, the axis representation is axis-order invariant, implying that it neither prioritizes one axis over another nor defines an object's long side. This attribute addresses the boundary discontinuities encountered in rotated bounding box-based methods, bolstering our method's resilience when handling objects with near-square or near-circular shapes. Consequently, this leads to more robust optimization and consistent prediction.
For effective optimization, we devise two specialized loss functions. 

\begin{itemize}
    \item \textbf{Max-Projection Loss}: This loss function is introduced to supervise point set learning and facilitate object convergence without explicit keypoint annotations. It encourages the model to learn the optimal set of points that best represent the shape and contours of the object. By focusing on the projection of points onto the object's boundary, this loss function helps the model converge toward a more accurate representation of the object's spatial extent.
    \item \textbf{Cross-Axis Loss}: Designed for robust axis representation learning, cross-axis loss transforms axis representation into label encoding. This is achieved by discretizing angles into bins and applying smoothing to enhance robustness. The resulting label encoding yields a cross-axis shape with four peaks, representing the primary directionalities of the object. We utilize cross-entropy loss to supervise this learning process.
\end{itemize}

Building on this flexible representation, we propose the Oriented-DETR model. This model extends the end-to-end detection Transformer framework with point-axis prediction. Specifically, our model introduces conditioned point queries and a points detection decoder to predict points. It leverages group self-attention to facilitate information interaction between point queries, enabling the capture of relationships between points and iterative refinement through multiple layers. Experimental results demonstrate effectiveness in main datasets, showing significant performance improvements in oriented object detection tasks. 

\section{Related Work}
\subsection{Oriented Object Detection}
Oriented object detection faces the challenge of accurately representing objects with arbitrary orientations, whose axes often do not align with the image axes. While numerous studies~\cite{roi,xie2021oriented,redet,arc,lsk} have emphasized feature extraction techniques for such objects, the representation of oriented objects remains a fundamental issue.
One common approach is the use of rotated bounding boxes. This representation has led to numerous research efforts aimed at enhancing its robustness. Some methods~\cite{yang2020arbitrary,yang2021dense,yu2023phase,gwd,kld} utilize $({x,y,w,h,\theta})$ to define each rotated bounding boxes. 
They focus on addressing the boundary issues caused by the periodicity of angles, such as CSL~\cite{yang2020arbitrary}, which resolves the angular periodicity problem by transforming angle regression into a classification problem. However, these methods overlook the fundamental flaws in definition boundaries, which also lead to discontinuous loss. Some methods use different descriptions. Gliding Vertex~\cite{gliding} and RSDet~\cite{rsdet} propose to describe objects using the outer horizontal bounding box along with the offsets of the four vertices. However, They also encounter the issue of vertices regression order. Oriented Reppoints~\cite{Li_2022_CVPR} and OSKDet~\cite{oskdet} introduce point representation for oriented objects. Although avoiding the issue of abrupt definition changes, these point-based methods struggle to accurately describe both the position and orientation of shape-irregular objects. STD\cite{STD} employs a strategy of decoupling spatial transforms sequentially while CrackDet~\cite{CrackDet} partitioning the 180-degree regression task into four sub-tasks. Compared to these methods, our Point-Axis representation introduces, for the first time a decoupled representation of position and orientation for oriented object detection.

\subsection{Detection Transformers}
The Transformer framework~\cite{Transformer} has emerged as a promising approach for various computer vision tasks~\cite{VIT,wei2021scene,PETR,ARTrack,ARTrackv2}, especially for object detection~\cite{carion2020end,zhu2020deformable,liu2022dab, SAM-DETR, EfficientMS}. Compared to traditional CNN-based detectors~\cite{girshick2015fast,cascade,dynamic,tian2019fcos,wang2023yolov7}, DETR offers a simpler design, end-to-end optimization, and improved accuracy and scalability~\cite{zhang2022dino,zong2023detrs,groundingdino,maskdino,visua-incontextl}. Several studies~\cite{emo2,ao2,zeng2024ars} have attempted to adapt the DETR framework for oriented object detection. Typically, these methods model object queries as rotated bounding boxes and refine them using a transformer decoder. However, they neglect the fact that the iterative updating of queries by the previous DETR decoder~\cite{liu2022dab,zhang2022dino} is based on the premise that horizontal boxes are aligned with the image axes. Furthermore, they suffer from the previously discussed  representation issues of oriented objects. In this study, we seamlessly integrate the point-axis representation with the DETR framework to resolve the aforementioned issues.

\begin{figure*}[t]
	\centering
	\includegraphics[width=0.99\linewidth]{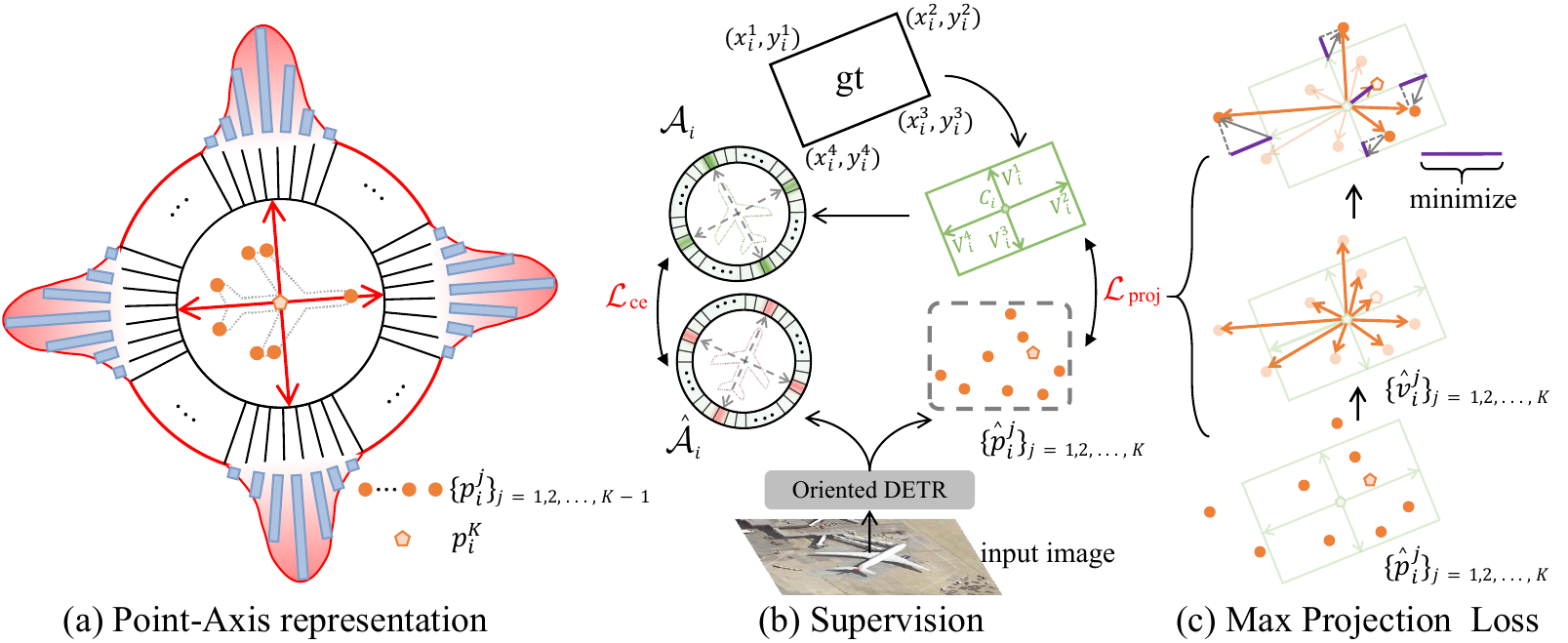}
	\caption{The overall framework of the point-axis representation. (a) provides a visual depiction of the representation method. (b) outlines the overall process of loss constraints. (c) introduces the key loss, max projection loss, proposed by us for supervising the point set learning. }
	\label{fig:representation}
\end{figure*}

\section{Oriented DETR with Point-Axis Representation}
In this section, we introduce the point-axis representation and elaborate on its implementation, along with the corresponding loss constraints. Using this representation as a foundation, we present the Oriented DETR model and detail its architectural framework, emphasizing the design of the point-axis prediction.

\subsection{Point-Axis Representation}

We propose the point-axis representation, as illustrated in Figure~\ref{fig:representation} (a). In this representation, each oriented object $i$ is defined by a combination of a set of points $\mathcal{P}_i$ and an axis representation $\mathcal{A}_i$. The set of points, denoted as $\mathcal{P}_i=\{ p_i^j \}_{j=1,2,\dots, K}$, comprises $K$ points, with the $K$-th point designated as the object's center. For the axis representation, we discretize directions into bins and apply Gaussian smoothing to generate a four-peak label encoding that represents the primary directionalities of the object. Notably, our representation exhibits the following advantages:

\begin{itemize}
    \item The decoupling of orientation from the bounding box prevents abrupt changes at boundaries that occur due to the coupled definition in methods based on rotated bounding boxes. This separation ensures a smoother transition between orientations, enhancing the stability and reliability of the representation for oriented objects.
    \item Employing circular labels ensures the consistency of angles at both the beginning and end of the defined range, thereby avoiding abrupt losses caused by angle periodicity. For instance, labels for 0$^{\circ}$ and 360$^{\circ}$ are identical, minimizing loss between 0$^{\circ}$ and 359$^{\circ}$. This approach maintains continuity and smoothness in representing angles, enhancing the robustness and effectiveness of the orientation encoding process.
\end{itemize}
The point-axis representation provides a detailed depiction of oriented objects by incorporating positional information via point sets and directional details through axes. In the following sections, we delve into the loss constraints associated with this representation, elucidating how they contribute to the overall effectiveness of our approach.

{\flushleft{\textbf{Point-Axis Loss Functions.}}}
Toward effective model optimization, we propose two specialized loss functions as depicted in Figure~\ref{fig:representation} (b). For each object $i$, we initially transform its ground truth bounding box, defined by four coordinates $\{({x}_i^j, {y}_i^j)\}_{j=1,2,3,4}$, into a central point $C_i=\{ (x_i^c, y_i^c)\}$, accompanied by radial vectors extending from the central point to the edges of the bounding boxes $\mathcal{V}_i=\{ v_i^j \}_{j=1,2,3,4}$. From these vectors, we derive $\mathcal{A}_i$. Subsequently, our model generates a set of points $\mathcal{\hat{P}}_i$ and predicted axis encoding $\mathcal{\hat{A}}_i$. Finally, we enforce constraints on these predictions using max-projection loss and cross-axis loss.

The overall Point-Axis loss function is defined as:
\begin{equation}
    \mathcal{L} = \frac{1}{N}\sum_{i=1}^{N}(\lambda_1\mathcal{L}_{proj}(\hat{\mathcal{P}}_i,\mathcal{V}_i, C_i)+ \lambda_2 \mathcal{L}_{ca}(\hat{\mathcal{A}}_i,\mathcal{A}_i)),
\end{equation}
where $N$ is the total number of instances, $\lambda_1$ and $\lambda_2$ are balancing coefficients, $\mathcal{L}_{proj}$ denotes our proposed max-projection loss, and $\mathcal{L}_{ca}$ represents the cross-axis loss. We will now detail each loss function.

{\flushleft{\textbf{Max-Projection Loss.}}}
The max-projection loss is concise and easy to implement, with the computation process illustrated in Figure~\ref{fig:representation} (c). For the $i$-th object with its predicted point set $\mathcal{\hat{P}}_i$, we first convert it into a vector representation relative to $C_i$: $\mathcal{\hat{V}}_i = \{\hat{v}_i^j \}_{j=1,2,\dots,K}$. We then project each vector onto the elements of $\mathcal{V}_i$, selecting the element with the maximum projection value for optimization. The optimization objective is intuitively depicted in Figure~\ref{fig:representation} (c) and mathematically expressed as:
\begin{equation}
\text{minimize} \sum_{j=1}^{4} \left| \max_{\resizebox{0.5\width}{!}{{m=1,2,\dots,K-1}}} \left  (   \resizebox{4.0\height}{!}{$\frac{(\hat{v}_i^m - v_i^j) \cdot v_i^j}{\| v_i^j \|}$}   \right) \right| + \| \hat{v}_i^K \|.
\end{equation}
It is worth noting that the max-projection loss does not impose auxiliary constraints on non-projected local maxima in the set, which reduces the ambiguity in optimization direction and enhances the flexibility of point set description.

{\flushleft{\textbf{Cross-Axis Loss.}}}
The cross-axis loss is implemented using the cross-entropy loss, the mathematical
expression of the optimization objective is  as follows:
\begin{equation}
\text{minimize} \;\frac{1}{N_{bins}}\sum_{j=1}^{N_{bins}} \left[\mathcal{A}_i^j\log\hat{\mathcal{A}}{_i^j} + (1-\mathcal{A}_i^j)\log(1-\hat{\mathcal{A}}{_i^j}) \right].
\end{equation}
where N$_{bins}$ is the number of bins for discretizing directions, with a default setting of 360. During inference, we take the argmax of these bins to obtain the principal direction, and then expand to the other three directions at intervals of 90 degrees.

\subsection{Architecture}
\begin{figure*}[t]
	\centering
	\includegraphics[width=0.99\linewidth]{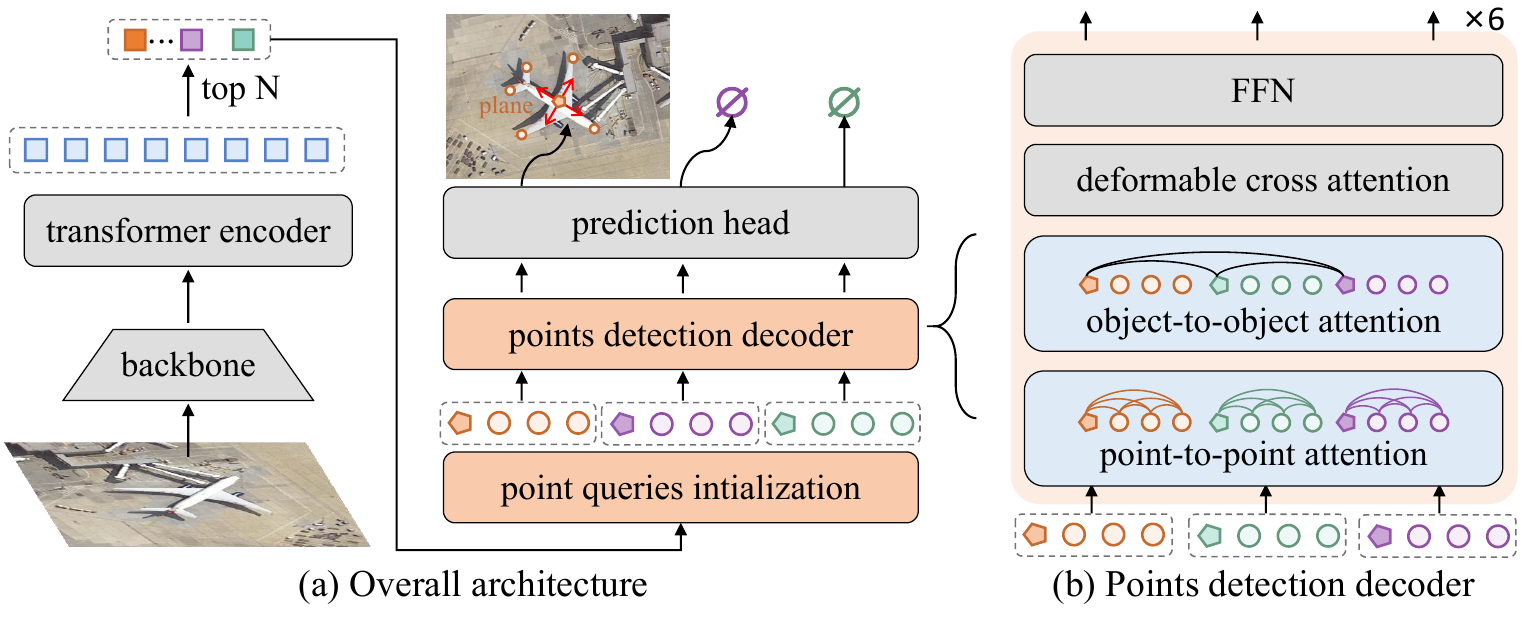}
	\caption{The architecture of Oriented DETR. }
	\label{fig:overallnetwork}
\end{figure*}
The model architecture of Oriented DETR is depicted in Figure~\ref{fig:overallnetwork}. 
Following the typical DETR framework, given an input image ${x}_{\rm img} \in \mathbb{R}^{\rm 3\times H \times W}$, it initially extracts features using a backbone and enriches them with global context through a transformer encoder. Subsequently, based on the scores corresponding to the features, it selects the top $N$ as object queries ${Q}_{o} \in \mathbb{R}^{N\times dim}$, where $N$ is the pre-defined maximum instance numbers. Our key design involves initializing point queries using an object-to-point query conversion function and enabling interaction among point queries using a points detection decoder. Finally, a prediction head is employed to generate the output.

{\flushleft{\textbf{Object-to-Point Query Conversion.}} }
For the $i$-th object query ${Q}_{o}^i$ with its reference point $(x_{ref}^i, y_{ref}^i)$, we convert it into $K$ point queries $\{ Q_p^{(i,j)}\}_{j=1,2,\dots,K}$ that are conditioned on ${Q}_{o}^i$. Among these, $Q_p^{(i,K)}$ is defined as the center point query corresponding to the definition of $\mathcal{P}_i$.
Following the common practice in DETR frameworks, we decompose the query into positional embeddings and content embeddings. For the center point query $Q_p^{(i,K)}$, we predict the offsets $(\Delta x^i, \Delta y^i)$ of the center point relative to the reference point $(x_{ref}^i, y_{ref}^i)$ using a multi-layer perceptron (MLP) layer:
\begin{equation}
\{\Delta x^i, \Delta y^i\} = MLP({Q}_{o}^i).
\end{equation}
The positional embedding $P_i^K$ for the center point query is then generated by concatenating the positional encodings of the updated center point coordinates:
\begin{equation}
{P_i^K} = Concat(PE(x_{ref}^i + \Delta x^i), PE(y_{ref}^i + \Delta y^i)),
\end{equation}
where $PE$ is a positional encoding function.

For the other point queries, we generate their positional embeddings based on the polar coordinates. Initially, we define $K-1$ equidistant angles ${\{ \theta_j\}}_{j=1,2,\dots,K-1}$ and establish a polar coordinate system with $(x_{ref}^i, y_{ref}^i)$ 
as the origin. Subsequently, we utilize an MLP layer to predict the polar coordinates of the object boundary in these directions and generate $P_i^{j}$ as follows:
\begin{equation}
\begin{aligned}
{\{\Delta r_j^i\}}_{j=1,2,\dots, K-1} =& MLP({Q}_{o}^i),\\
P_i^j = Concat(PE(x_{ref}^i+ \Delta r_j^i \cdot \cos{\theta_j}& ),PE(y_{ref}^i+ \Delta r_j^i \cdot \sin{\theta_j})).
\end{aligned}
\end{equation}
Finally, we use ${Q}_{o}^i$ as the content embedding for all point queries to construct the initial point queries conditioned on $i$-th instance. 

{\flushleft{\textbf{Points Detection Decoder.}} }
In the DETR framework, the decoder plays a pivotal role in facilitating information exchange between different components. It typically consists of self-attention mechanisms to enable interaction among queries and cross-attention mechanisms to promote interaction between queries and image features. Oriented DETR extends this concept by introducing a points detection decoder specifically designed to handle point queries. The point detection decoder in Oriented DETR comprises two key modules: the point-to-point attention module and the object-to-object attention module, as illustrated in Figure~\ref{fig:overallnetwork} (b).

In the point-to-point attention module, we divide point queries into $N$ groups according to their corresponding instances, and each group contains $K$ point queries. Then we apply a parameter-shared self-attention layer within each group to capture the relationships between points. In object-to-object attention, we extract the center point query from each instance to form a new group $G_c=\{Q_{p}^{(i,K)}\}_{i=1,2,\dots, N}$ and apply self-attention within this new group to capture relationships between objects. Afterward, all queries are passed through a deformable cross-attention module and a $FFN$ module for further processing.

{\flushleft \textbf{Prediction Head.}}
In the prediction head, for each instance $i$, each point query $Q_i^j$ is mapped to a 2D point coordinate $\hat{p}_i^j$, while its class $\hat{c_i}$ and axis $\hat{\mathcal{A}_i}$ are mapped from all its conditioned point queries $\{ Q_p^{(i,j)}\}_{j=1,2,\dots,K}$.

\section{Experiment}
\subsection{Evaluation Datasets}
{\flushleft \textbf{DOTA}~\cite{dota}} is a large oriented object detection dataset comprising 2,806 images and 188,282 instances distributed across 15 categories, with a substantial variation in orientations, shapes, and scales. The image sizes range from 800$\times$800 to 4000$\times$4000. The split ratio for training, validation, and testing sets is 3:1:2. According to the official toolkit, DOTA has two evaluation protocols: 1) single-scale training and testing 2) multi-scale training and testing. For single-scale training and testing, the training and testing images are cropped into 1024$\times1024$ patches with a stride of 824. For multi-scale training and testing, the raw pictures are first resized at three scales (0.5, 1.0, and 1.5) and then cropped into 1024$\times$1024 patches with a stride of 524. 

{\flushleft \textbf{DIOR-R}~\cite{dior}} is constructed based on DIOR~\cite{li2020object}, with the addition of annotations for rotated bounding boxes. It comprises 23,463 images with a size of 800$\times$800 pixels, totaling 190,288 oriented instances across 20 categories. The dataset is divided into training, validation, and testing sets with a ratio of 1:1:2.

{\flushleft \textbf{HRSC2016}~\cite{liu2017high}} is a renowned dataset for ship detection, comprising 1,061 images collected from ports with sizes ranging from 300$\times$300 to 1500$\times$900. The training, validation, and testing sets consist of 436, 181 and 444 images, respectively. 

In addition to the above datasets for oriented object detection, we also extend our model to handle general object detection tasks. To this end, we evaluate our approach on the \textbf{COCO2017}~\cite{lin2014microsoft} dataset, which is a widely used benchmark for general object detection. It allows us to demonstrate the versatility and generalizability of our proposed approach.

\subsection{Implementation Details}
When comparing with other methods on DOTA, DIOR-R, and HSRC2016, we use training and validation set for training and testing set for evaluation. The extension experiments on COCO are conducted by training on the training set and evaluating on the validation set. For DOTA, DIOR, and COCO, we train the model for 36 epochs, with a learning rate reduction at epochs 27 and 33. For HSRC2016, we train the model for 50 epochs, with learning rate reductions at epochs 33 and 45. We use the AdamW optimizer with an initial learning rate of 0.0001 and weight decay of 0.0001. Training is performed on 4 RTX 4090 GPUs with a total batch size of 8. Inference is performed on a single RTX 4090 GPU. The complete hyperparameters can be found in the supplementary file.

\begin{figure}[t]
    \centering\small
    \begin{minipage}[t!]{0.74\textwidth}
        \centering\small
        \includegraphics[width=0.99\columnwidth]{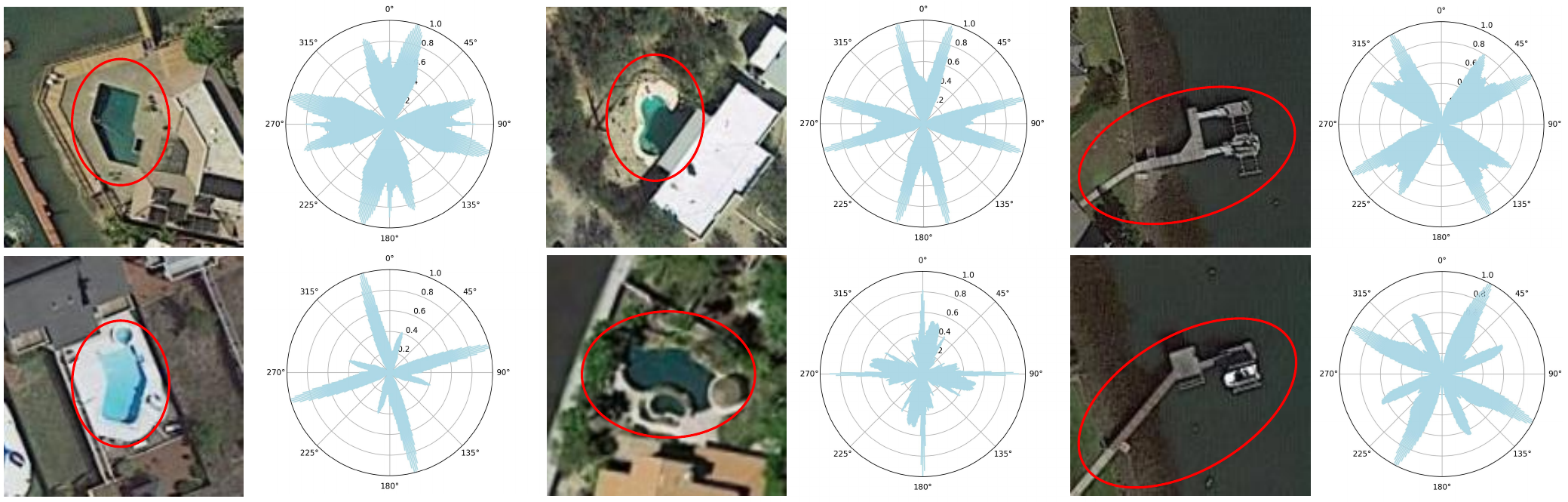}
    \end{minipage}
    \hskip0.5em
    \begin{minipage}[t!]{0.22\textwidth}
        \caption{The distributions of predicted axes for objects without clear orientation definition.}
        \label{fig:swim_distibution_vis}
    \end{minipage}
\end{figure}

\begin{figure*}[h]
	\centering
	\includegraphics[width=0.98\linewidth]{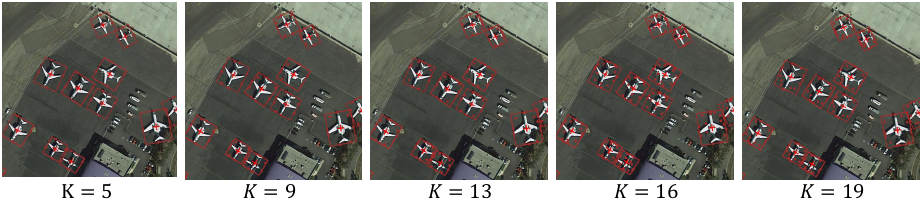}
	\caption{Visualization of results under different numbers of points.}
    \label{point_number}
\end{figure*}

\subsection{Ablation Study and Analyses}

The key components of Oriented DETR are the point-axis representation and the points detection decoder. In our ablation study, we analyze these crucial components with ResNet-50 as backbone, training on the training set and evaluate on the evaluating on the validation set. Default settings are marked in \colorbox{gray!25}{gray}.

{\flushleft \textbf{Axis Representation Learning.}} 
To understand the behavior of axis predictions, we visualize their distributions for various objects in the DOTA dataset. As shown in Figure~\ref{intro_vis}, for most objects, the axis predictions accurately capture the directional information. However, we also identify some anomalous distributions. Upon closer inspection, these anomalies often correspond to objects that lack an overall clear directional definition, such as certain swimming pools, as illustrated in Figure~\ref{fig:swim_distibution_vis}. It's worth noting that even in the DOTA dataset, annotations for such objects exhibit ambiguity. Nevertheless, our model is capable of learning the directional information for these objects and providing predictions that cover all possible directions through distributions. This analysis highlights the robustness of our approach in handling objects with varying degrees of directional clarity and demonstrates its ability to provide meaningful predictions even in the presence of annotation ambiguities. Additional visualizations of predicted axis distributions can be found in the supplementary file.

\begin{table}[t]
\centering
\def\arraystretch{1.5}
\begin{minipage}{0.49\linewidth}\flushleft

\caption{Performance comparison \\with different numbers of points.}
\tiny
\setlength{\tabcolsep}{1.0mm}{
\begin{tabular}{c|ccccc}
        \toprule
         Values of $K$ &5 &9 & 13 & 16 &19  \\
         \hline
         mAP$_{50}$&74.86&74.93&\default{75.35}&\textbf{75.53} &75.10\\
         mAP$_{75}$&49.64&49.98&\default{\textbf{50.14}}&50.03 &49.60\\
         \bottomrule
        \end{tabular}
\label{tab:number_compare}
    }
\end{minipage}
\begin{minipage}{0.49\linewidth}\centering
\caption{Performance comparison \\with different point constraints.}
\tiny
\def\arraystretch{1.5}
\setlength{\tabcolsep}{0.8mm}{
 \begin{tabular}{c|c|c|c|c}
        \toprule
         \multirow{2}*{\scriptsize{Loss}} & \multirow{2}*{Max-Projection}&\multicolumn{3}{c}{Max-Projection variants}\\
         \cline{3-5}
           & & w.penalty&top-2\qquad&top-3\qquad\\
         \hline
         mAP$_{50}$&\default{\textbf{75.35}}&75.20&74.77&73.20\\ 
         mAP$_{75}$&\default{\textbf{50.14}}&50.02&49.36&47.88\\ 
         \bottomrule
        \end{tabular}
\label{tab:loss_compare}
    }
\end{minipage}
\end{table}

\begin{figure*}[t]
	\centering
	\includegraphics[width=0.95\linewidth]{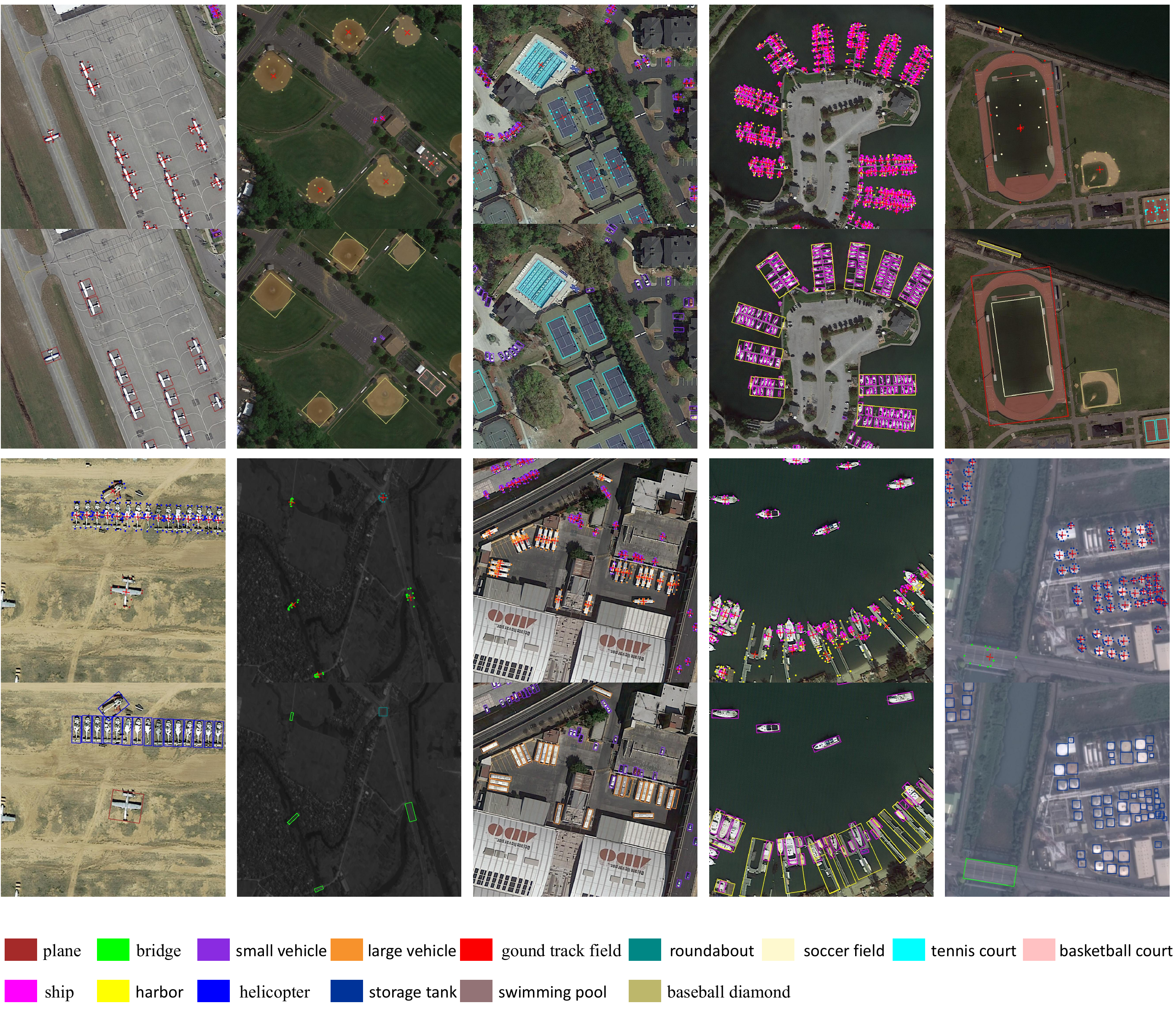}
	\caption{Visualization of results from Oriented DETR in DOTA.}
	\label{fig:visualization}
\end{figure*}

\begin{table}[t]
    \centering
    \renewcommand{\arraystretch}{1.2}
    \caption{Ablation study on the design of the Points Detection Decoder.}
    \scriptsize
    \setlength{\tabcolsep}{2.8mm}{
        \begin{tabular}{c|c|ccc}
        \toprule

           & Baseline & \multicolumn{3}{c}{Decoder components}\\
         \midrule
         Point Queries?& &\checkmark& \checkmark &\checkmark\\
         Group Self-Attention?& & &\checkmark&{\checkmark}\\
         Decouple Cross-Attention?& & & &{\checkmark}\\
         \hline
         mAP$_{50}$&72.80& 70.98& 74.21&\default{\textbf{75.35}} \textbf{($\uparrow$ 2.55)}\\
         mAP$_{75}$&45.25& 44.06& 48.30&\default{\textbf{50.14 ($\uparrow$ 4.89)}}\\
         \bottomrule
        \end{tabular}}
	    \label{tab:decoder ablation}
\end{table}

{\flushleft\textbf{Number of Points.}} 
We investigate the influence of $K$ (the number of points representing each object) on model performance. Table~\ref{tab:number_compare} and Figure~\ref{point_number} respectively present quantitative metrics and visual results. Statistically, increasing the number of points from 5 to 13 yielded a modest improvement of \textbf{0.49\%} for mAP$_{50}$ and \textbf{0.50\%} for mAP${_{75}}$. Visually, augmenting the number of points facilitates more precise localization. Nonetheless, even with only four boundary descriptor points at $K=5$, predictions can still be accurately translated into rotated bounding boxes, predominantly owing to the auxiliary axis predictions.

{\flushleft\textbf{Point Loss Constraints.}}
In our method, the max-projection loss only imposes constraints on the maximum projection value in each direction. We seek to explore whether incorporating additional constraints could enhance supervision. Initially, we experiment by calculating the loss for all points located outside the ground-truth box and incorporating this into the penalty term. Furthermore, we also attempt to constrain the top-$k$ values of the projection, rather than solely focusing on the maximum value. 
However, the experimental result in Table~\ref{tab:loss_compare} indicates that introducing additional penalties and increasing loss constraints don't result in improved accuracy. Specifically, when considering the top-$k$ variant, we observe a notable decrease in accuracy when $k=2$. We hypothesize that imposing excessive constraints may diminish the flexibility of the point set representation, ultimately leading to ambiguous optimization directions.

\vspace{-0.2cm}
{\flushleft \textbf{Points Detection Decoder.}} We conduct ablation experiments on different components of the points detection decoder, with the results shown in Table~\ref{tab:decoder ablation}. Starting from a two-stage deformable DETR~\cite{zhu2020deformable} (object queries in the decoder are generated from encoder output features) and retaining the object queries design, we utilize each object query to predict a set of points and an orientation axis, which serves as our baseline. This yields a mAP$_{50}$ of \textbf{72.80$\%$} and a mAP$_{75}$ of \textbf{45.25$\%$}. Subsequently, we model queries as dynamic points and use a vanilla self-attention layer to handle interactions between point queries, resulting in decreases of \textbf{1.82$\%$} for mAP$_{50}$ and \textbf{1.19$\%$}for mAP$_{75}$. We attribute this to the ambiguity in the interaction between point queries of different objects. Therefore, we adopt group self-attention, initially grouping point queries by instances and then performing self-attention operations within each group, which significantly improves detection accuracy. Further, we decouple the deformable cross-attention modules for center point queries and boundary point queries to facilitate specialized feature learning, further improving its performance. Ultimately, the design of the points detection decoder leads to increases in accuracy of \textbf{2.55$\%$} for mAP$_{50}$ and \textbf{4.89$\%$} for mAP$_{75}$.

\subsection{Comparison with State-of-the-Art Methods}

{\flushleft \textbf{Results on DOTA.}}
We firstly conduct a fair comparison with state-of-the-art methods using the single-scale training and testing protocol, as commonly employed in most papers. The results are presented in Table~\ref{table:dota_ss_compare}. Using ResNet50~\cite{resnet} and Swin-T~\cite{swin} as backbones, Oriented DETR achieved mAP$_{50}$ of \textbf{79.1$\%$} and \textbf{79.8$\%$} respectively, outperforming all previous methods. Using the same backbone, Oriented DETR outperforms the point-based representation method Oriented Reppoints by \textbf{2.2$\%$}. Compared to end-to-end DETR-based methods EMO2-DETR, ARS-DETR, and AO2-DETR, it achieves performance gains of \textbf{7.5$\%$}, \textbf{4.3$\%$}, and \textbf{1.4$\%$}, respectively. Table~\ref{tab:sys_compare} shows a systematic comparison (multi-scale training and testing) with others. Using Swin-L as the backbone, we achieve an mAP$_{50}$ of \textbf{82.26$\%$}, surpassing the previous leading method, LSKNet, by \textbf{0.41$\%$}, and outperforming the best DETR-based method, AO2-DETR, by \textbf{3.04$\%$}. The visualization results are shown in Figure~\ref{fig:visualization}.

{\flushleft \textbf{Results on DIOR.}}
Table~\ref{tab:dior_compare} shows our results on the DIOR dataset which has a richer variety of targets in terms of quantity and categories. using Swin-T as the backbone, Oriented DETR achieves an mAP{$_{50}$} of \textbf{74.26$\%$}, surpassing the previous best method by \textbf{3.21$\%$}. Using ResNet50 as the backbone, Oriented DETR achieves an mAP$_{50}$ of \textbf{66.80$\%$}, surpassing the best method with the same backbone by \textbf{2.89$\%$}.
{\flushleft \textbf{Results on HRSC2016.}}
On the HRSC2016 dataset, Oriented DETR achieves competitive results, as shown in Table~\ref{tab:hrsc_compare}. Oriented DETR achieves the highest accuracy of \textbf{98.02$\%$} under the PASCAL VOC 2007~\cite{voc07} metrics and achieves the second-best accuracy of \textbf{90.56$\%$} under the PASCAL VOC 2012~\cite{voc12} metrics, with a slight disadvantage compared to the previous best method (\textbf{90.60$\%$}). On the HRSC2016 dataset, objects tend to have large aspect ratios and simple shapes, making the boundary issues almost non-existent for methods based on rotated bounding boxes of long-edge definition. Despite not having a significant advantage in accuracy, we have demonstrated the generalization capability of Oriented DETR across different scenarios.

\begin{table}[t]
\setlength{\tabcolsep}{3pt}
\begin{center}
\caption{Comparison with state-of-the-art methods on DOTA under single scale training and testing protocol. Best in \textbf{bold}. }
\label{table:dota_ss_compare}
\resizebox{\linewidth}{!}{
\begin{tabular}{rccccccccccccccccc}

    \toprule
        \textbf{Methods}  & \textbf{Backbone}  & PL& BD & BR& GTF& SV& LV& SH& TC& BC & ST & SBF & RA & HA & SP & HC & {mAP$_{50}$} \\
    \midrule
    
        \multicolumn{18}{c}{\textbf{one-stage methods}}\\
        R$^3$Det~\cite{yang2021r3det} & R-152 & 89.5 & 81.2 &50.5 &66.1 &70.9 &78.7 &78.2 &90.8 &85.3 &84.2 &61.8 &63.8 &68.2 &69.8 &67.2 &73.7\\
         
        SASM~\cite{hou2022shape} & R-50 &86.4 &79.0 &52.5 &69.8 &77.3 &76.0 &86.7 &90.9& 82.6 &85.7 &60.1 &68.3 &74.0 &72.2 &62.4 &74.9\\

        GGHL~\cite{huang2022general}  &Darknet53  &89.7& 85.6 &44.5 &77.5 &76.7 &80.5 &86.2 &90.8 &88.2 &86.3 &67.1 &69.4 &73.4 &68.5 &70.1 &77.0 \\
        
        Orient-Rep~\cite{Li_2022_CVPR}& Swin-T  &  89.1 & 82.3 & 56.7 & 75.0 & 80.7 & 83.7& 87.7 & 90.8 & 87.1 & 85.9 & 63.6 & 68.6 & 75.9 & 73.5 & 63.8 & 77.6 \\  
        DHRec~\cite{9832808} & R-101 &88.8 &82.1 &55.7 &72.0 &77.8 &79.1 &88.0 &90.9 &84.0 &85.8 &58.0 &66.6 &74.5 &70.6 &57.3 &75.4\\
    
        DCFL~\cite{xu2023dynamic} & ReR101 &- &- &- &- &- &- &- &- &- &- &- &-  &- &- &- &75.4 \\
        
        PSC~\cite{yu2023phase} & Darknet53 &89.9 &86.0 &54.9 &62.0 &81.9 &85.5 &88.4 &90.7 &86.9 &88.8 &63.9 & 69.2 &76.8 &82.8 &63.2 &\underline{78.1} \\

        \midrule
        \multicolumn{18}{c}{\textbf{two-stage methods}}\\

        SCRDet~\cite{yang2019scrdet} & R-101 & 90.0	&80.6	&52.1	&68.4	&68.4	&60.3	&72.4	& 90.9  &87.9	&86.9	&65.0 &66.7	&66.3	&68.2	&65.2	&72.6 \\
        RoI-Trans~\cite{roi} & R-101 &88.6& 82.6 &52.5& 70.9 &77.9& 76.7& 86.9 &90.7 &83.8& 82.5 &53.9 &67.6 &74.7& 68.8 &61.0 &74.6\\
        
        G. Vertex~\cite{gliding} & R-101 & 89.6 &85.0 &52.3 & 77.3 &73.0 &73.1 & 86.8 &90.7 &79.0 & 86.8 & 59.6 &70.9 &72.9 & 70.9 & 57.3 & 75.0 \\ 

        CSL ~\cite{yang2020arbitrary} & R-152 & 90.3 &85.5 &54.6 &75.3 &70.4 &73.5 &77.6 &90.8 &86.2 &86.7 &69.6 &68.0 &73.8 &71.1 &68.9 &76.2\\
        
        O-RCNN~\cite{xie2021oriented} & R-101 &88.9& 83.5 &55.3& 76.9 &74.3& 82.1& 87.5 &90.9 &85.6& 85.3 &65.5 &66.8 &74.4& 70.2 &57.3 &76.3\\

        ReDet ~\cite{redet}  &ReR50 &88.8 &82.6 &54.0 &74.0 &78.1 &84.1 &88.0 &90.9 &87.8 & 85.8 &61.8 &60.4 &76.0 &68.1 &63.6 &76.3\\

        OSKDet~\cite{oskdet} & R-101 &90.1& 87.1 &54.2& 75.6 &72.6& 76.9& 87.6 &90.8 &79.1 & 86.9 &59.9 &71.3 &75.2& 71.7 &66.7 &76.4\\

        RVSA~\cite{wang2022advancing} & ViTAE-B &89.4 &84.3 &59.4 &73.2 &80.0 &85.4 &88.1 &90.9 &88.5 &86.5 &58.9 &72.2 &77.3 &79.6 &71.2 &\underline{79.0}\\
        
        ARC~\cite{arc} & ARC-R101 &89.4 &83.6 &57.5 &75.9 &78.8 &83.6 &88.1 &90.9 &85.9 &85.4 &64.0 &68.7 &75.6 &72.0 &65.7 &77.7\\
        COBB~\cite{COBB} & ResNet50 &- &- &- &- &- &- &- &- &- &- &- &-  &- &- &- &76.6\\
        PKINet~\cite{pki} &PKINet-S &89.7 &84.2 &55.8 &77.6 &80.3 &84.5 &88.1 &90.9 &87.6 &86.1 &66.9 &70.2 &77.5 &73.6 &62.9 &78.4\\

        \hline\noalign{\smallskip}
        \multicolumn{18}{c}{\textbf{DETR based end-to-end methods}}\\
        \noalign{\smallskip}
        AO2-DETR~\cite{ao2} & R-50 &89.3 &85.0 &56.7 &74.9 &78.9 &82.7 &87.4 &90.5 &84.7 &85.4 &62.0 &70.0& 74.7& 72.4& 71.6& \underline{77.7}\\
        EMO2-DETR~\cite{emo2} & Swin-T &89.0& 79.6 &48.7& 60.2 &77.3 & 76.4& 84.5 &90.8 &84.8 & 85.7 &48.9 &67.6 &66.3& 71.5 &53.5 &72.3\\
        ARS-DETR~\cite{zeng2024ars} & Swin-T &87.7 &76.5 &50.6 &69.9 &79.8 &83.9 &87.9 &90.3 &86.2 &85.1 &54.6 &67.0 &75.6 &73.7 &63.4 &75.5 \\
        \midrule
        \textbf{Our Oriented-DETR}& R-50 &89.2 & 86.4	&57.7	&75.3	&81.1	&84.7	&89.1	&90.9	&86.1	&87.0	&59.5	&70.3	&79.3	&81.5	&68.8	&79.1	\\
        \textbf{Our Oriented-DETR}  & Swin-T &89.4  &85.1 &57.8 &75.0 &81.2 &86.1 &89.1 &90.9 &88.7 &87.0& 62.9& 69.1& 80.7& 82.8& 71.0 & \textbf{79.8}\\
    \bottomrule
\end{tabular}
}
\end{center}
\vspace{-10pt}
\end{table}
\setlength{\tabcolsep}{1.5pt}

\begin{table}[t]
    \centering
    \caption{System comparison with state-of-the-art methods on DOTA.}
    \label{tab:sys_compare}
    \setlength{\tabcolsep}{6pt}
    \resizebox{\linewidth}{!}{
        \begin{tabular}{crrrrrrr}
        \toprule
         Methods& EMO2-DETR~\cite{emo2} & SASM~\cite{hou2022shape} & AO2-DETR~\cite{ao2} & ReDet~\cite{redet}& R$^3$Det-KLD~\cite{kld} & AOPG~\cite{cheng2022anchor}& O-RCNN~\cite{xie2021oriented}\\
         mAP$_{50}$& 78.46 & 79.17 & 79.22 & 80.10 & 80.63 &80.66 &80.87 \\
         \midrule
         Methods & KFIoU~\cite{yang2022kfiou} & RVSA~\cite{wang2022advancing} & RTMDet-L~\cite{lyu2022rtmdet} & ARC~\cite{arc} & LSKNet-S~\cite{lsk} &\textbf{Ours(Swin-T)}&\textbf{Ours(Swin-L)}\\

         
         mAP$_{50}$ & 80.93 & 81.24 &81.33 &81.77 & 81.85 & 81.78 & \textbf{82.26}\\
         \bottomrule
        \end{tabular}}
        \vspace{-10pt}
\end{table}

\begin{table}[t]
\centering
\def\arraystretch{1.0}
\begin{minipage}{0.49\linewidth}\flushleft

\caption{Performance comparisons on \protect\\the DIOR-R dataset.}
\label{tab:dior_compare}
\tiny
\setlength{\tabcolsep}{2.0mm}{

\begin{tabular}{rcc}
        \toprule
         Methods & Backbone & mAP$_{50}$\\
         \midrule
         Gliding Vertex~\cite{gliding}& R50 &60.06\\ 
         TIOE-Det~\cite{ming2023task} & R50 & 61.98\\
         ROI-Trans~\cite{roi} & R50 & 63.87\\ 
         AOPG~\cite{cheng2022anchor} & R50 & 64.41\\
         O-RCNN~\cite{xie2021oriented} & R50 &64.30\\
         ARS-DETR~\cite{zeng2024ars} & R50 & 66.12\\
         AO2-DETR~\cite{ao2} & R50 & 66.41\\
         GGHL~\cite{huang2022general} &R50 &66.48\\
         Oriented RepPoints~\cite{Li_2022_CVPR} & R50 &66.71\\
         DCFL~\cite{xu2023dynamic} & R50& 66.80\\
         DCFL~\cite{xu2023dynamic} & ReR101& 71.03\\
         RVSA~\cite{wang2022advancing} & ViTAE-B & 71.05\\
         \midrule
         \textbf{Our Oriented-DETR} & R50 & 69.69\\
         \textbf{Our Oriented-DETR} & Swin-T & \textbf{74.26}\\
         \bottomrule
        \end{tabular}
    }
    
\end{minipage}
\begin{minipage}{0.49\linewidth}\flushleft
\caption{Performance comparisons on the HSRC2016 dataset.}
\label{tab:hrsc_compare}
\tiny
\setlength{\tabcolsep}{0.8mm}{
\begin{tabular}{rccc}
        \toprule
         Methods & Backbone & mAP(07) & mAP(12)\\
         \midrule
         ROI-Trans~\cite{roi} & R101 & 86.20 &-\\ 
         AO2-DETR~\cite{ao2} & R50 & 88.12 &97.47\\
         R$^3$Det~\cite{yang2021r3det} & R101 &89.26&96.01\\
         GWD~\cite{yang2022detecting} & R101 &89.85 &97.37\\
         OSKDet ~\cite{oskdet} & R101 & 89.98 &-\\
         PSC~\cite{yu2023phase} &DarkNet53 & 90.06  & -\\
         AOPG~\cite{cheng2022anchor} & R101 &90.34 &96.22\\
         Oriented Reppoints~\cite{Li_2022_CVPR} & R50 & 90.38 & 97.26\\
          ARC~\cite{arc} &ARC-R50 & 90.41  & -\\
          ReDet~\cite{redet} &ReR50 &90.46 &97.63\\
         Oriented-RCNN~\cite{xie2021oriented} & R101 &90.50 &97.60\\
          RTMDet~\cite{lyu2022rtmdet} & RTM-L &  \textbf{90.60} &97.10\\
          \midrule
         \textbf{Our Oriented-DETR} & R50 & 90.52 & 97.73 \\
         \textbf{Our Oriented-DETR} & Swin-T &90.56 & \textbf{98.02}\\
         \bottomrule
        \end{tabular}
    }
\end{minipage}
\end{table}

\subsection{Extension to General Object Detection Task}
We extend Oriented DETR to general object detection which only needs to output horizontal detection boxes and use COCO for evaluation. Concretely, we consider the axes of objects to be fixed in the horizontal and vertical directions, and remove the prediction and supervision for axes. Table~\ref{table:coco} lists the experimental results, showing that Oriented DETR still maintains a precision advantage over other DETR baseline methods. 
Future work could explore combining Oriented DETR with more advanced training strategies, such as those proposed in~\cite{li2022dn,zhang2022dino,zong2023detrs,jia2023detrs}, to further enhance its performance and versatility.

\setlength{\tabcolsep}{4.0pt}
\begin{table}[t]
\renewcommand{\arraystretch}{1.0}
\scriptsize
\begin{center}
\caption{Results on COCO ${val}$.}
\label{table:coco}
\resizebox{\linewidth}{!}{
\begin{tabular}{rcccccccc}
    
    \toprule
        \textbf{Methods} & Backbone & Epoches  & AP & $\rm AP_{50}$ & $\rm AP_{75}$& $\rm AP_S$& 
        $\rm AP_M$ & $\rm AP_L$\\
    \midrule
        \multicolumn{9}{c}{\textbf{baseline methods}}\\
        DETR~\cite{carion2020end}&R-50& 500 &43.3 & 63.1 &45.9 &22.5 &47.3 &61.1\\
        Deformable-DETR~\cite{zhu2020deformable} &R-50&50 &43.8 &62.6 &47.7 &26.4 &47.1 &58.0\\
        Deformable-DETR$^\dag$~\cite{zhu2020deformable} &R-50&50 &46.2 &65.2 &50.0 &28.8 &49.2 &61.7\\
        DAB-DETR~\cite{liu2022dab}&R-50& 50& 45.7&66.2&49.0&26.1&49.4&63.1\\
        DAB-Deformable-DETR$^{\dag\ddag}$~\cite{liu2022dab}&R-50&50&49.7 &68.2 &54.3 &32.0 &52.9 &65.3\\
        \textbf{Our Oriented-DETR} & R-50 & 36 &\textbf{50.5} & 68.1 &55.1 &33.2&53.5&65.0\\
        \midrule
        \multicolumn{9}{c}{\textbf{methods based on training strategies}}\\
        \rowcolor{gray!30}
        DN-Deformable-DETR ~\cite{li2022dn}& R-50&50& 48.6 &67.4 &52.7 &31.0 &52.0 &63.7\\
        \rowcolor{gray!30}
        $\mathcal{H}$-Deformable-DETR~\cite{zong2023detrs}&R-50&36 &50.0&-&-& 32.9 &52.7 &65.3\\
        \rowcolor{gray!30}
        DINO-4scale~\cite{zhang2022dino} & R-50& 36 &50.9& 69.0 &55.3& 34.6 & 54.1&64.6\\
        \rowcolor{gray!30}
        Group-DINO-4scale~\cite{jia2023detrs} & R-50&36 &51.3&-&-  &34.7 &54.5 &65.3 \\
    \bottomrule
    \multicolumn{9}{l}{$^\dag$: two-stage.}\\
    \multicolumn{9}{l}{ $^\ddag$: better result from detrex~\cite{detrex}.}
\end{tabular}
}
\end{center}
\end{table}
\setlength{\tabcolsep}{1.5pt}

\section{Conclusions}
This paper proposes the point-axis representation, which provides a stable representation for aerial-oriented object detection by decoupling the location and orientation information, accompanied by the point-axis loss functions to supervise both the point set and the axis without introducing additional annotations. Leveraging this representation, we present the Oriented DETR model which integrates the DETR framework for precise point-axis prediction and end-to-end detection, showing significant performance improvements in aerial-oriented object detection tasks. In the future, we hope this representation could be extended to detecting arbitrary polygonal bounding boxes. The code will be made publicly available to facilitate easy reproduction.

\section*{Acknowledgements}
This work was supported by the National Natural Science Foundation of China under Grant No. U21B2048 and No. 62302382, Shenzhen Key Technical Projects under Grant CJGJZD2022051714160501, the Fundamental Research Funds for the Central Universities No. xxj032023020, and sponsored by the CAAI-MindSpore Open Fund, developed on OpenI Community.

%
%
\bibliography{main}
\end{document}